# Intelligent Understanding of Large Language Models in Traditional Chinese Medicine Based on Prompt Engineering Framework

Yirui Chen, Qinyu Xiao, Jia Yi, Jing Chen, Mengyang Wang*

College of Public Health and Health Sciences
Tianjin University of Traditional Chinese Medicine
Tianjin, China

## Abstract

This paper explores the application of prompt engineering to enhance the performance of large language models (LLMs) in the domain of Traditional Chinese Medicine (TCM). We propose TCM-Prompt, a framework that integrates various pre-trained language models (PLMs), templates, tokenization, and verbalization methods, allowing researchers to easily construct and fine-tune models for specific TCM-related tasks. We conducted experiments on disease classification, syndrome identification, herbal medicine recommendation, and general NLP tasks, demonstrating the effectiveness and superiority of our approach compared to baseline methods. Our findings suggest that prompt engineering is a promising technique for improving the performance of LLMs in specialized domains like TCM, with potential applications in digitalization, modernization, and personalized medicine.

## 1 Introduction

In recent years, Traditional Chinese Medicine (TCM) has garnered significant interest due to its potential therapeutic advantages in the treatment of a multitude of diseases. Guo et al. (2020) underscored the utility of omics strategies for elucidating the therapeutic mechanisms of TCM at the molecular level. This investigative approach yields insights into the efficacy of TCM in addressing various diseases through intricate molecular interactions. Nevertheless, the establishment of stringent quality standards for TCM is imperative to guarantee its safety and therapeutic efficacy in clinical applications. Leong et al. (2020) conducted a comparative analysis of TCM quality standards between the European Pharmacopoeia and the Chinese Pharmacopoeia, underscoring the critical role of quality control in TCM to preclude medical adverse events.

In the domain of large language models (LLMs), recent breakthroughs have revealed their versatility across diverse applications. Huang et al. (2022) demonstrated the capacity of LLMs to refine their performance through the utilization of unlabeled datasets, exemplifying their adaptive and learning proficiency. Miao et al. (2023) introduced SpecInfer, a system designed to expedite the inference process of generative LLMs, thereby reducing latency and computational demands without compromising model integrity. Additionally, Peng et al. (2023) proposed the LLM-Augmenter system, which leverages external knowledge and automated feedback to enhance the responsiveness of LLMs, thereby improving overall model performance. The integration of LLMs with other modalities has also been investigated to broaden their functional scope.

Chen et al. (2023) introduced X-LLM, an innovation that conceptualizes multi-modalities as foreign languages, thereby equipping LLMs with multimodal capabilities. This strategy paves the way for LLMs to effectively process imagery, speech, and video content. Furthermore, Liu et al. (2023) presented LLM+P, a framework that amalgamates classical planning with LLMs, harnessing the strengths of both methodologies to optimize planning capabilities.

Within the TCM domain, researchers are focus on the application of LLMs to augment knowledge retrieval and reasoning. Hua et al. (2023) developed a specialized large language model attuned to TCM knowledge, enhancing clinical reasoning tasks such as diagnosis and treatment suggestions. Yizhen et al. (2024) examined the TCM knowledge comprehension of ChatGPT, highlighting the significance of assessing LLM performance within specialized fields like TCM. The development of TCM-specific LLMs has become a focal point of contemporary research. Zhang et al. (2024) introduced Qibo, a large language model tailored for TCM applications, which addresses challenges such as the scarcity of specialized corpus resources and the issue of overconfidence in predictions. Moreover, Yue et al. (2024) introduced TCMBench, a comprehensive benchmarking tool for evaluating the performance of LLMs in TCM-related tasks, providing a standardized framework for assessing their efficacy.In the rapidly advancing field of artificial intelligence, large models have exhibited exceptional prowess in the processing and comprehension of complex data. Traditional Chinese Medicine (TCM), with its rich historical legacy and extensive knowledge repository, is poised to gain substantial benefits from the integration of these sophisticated technologies. The current research on the intelligent parsing of large models in TCM, grounded in prompt engineering, is aimed at bridging the divide between the ancient insights of TCM and the forefront of machine learning techniques. This study endeavors to augment the interpretive and predictive capacities of large models through the application of prompt engineering—a pioneering method for guiding and refining model responses within the TCM domain. In doing so, our objectives extend beyond the digitalization and modernization of TCM practices to include contributions to the advancement of personalized and precision medicine, firmly rooted in the tenets of TCM.

## 2 Methods

As delineated in Section 1, prompt-based learning constitutes a synthesis of pre-trained language models (PLMs), human-elicited knowledge, and specialized natural language processing (NLP) tasks. Adhering to this conceptual framework, the design ethos is centered on the dual objectives of modular independence and interconnectivity.In the subsequent discussion, we commence by elucidating the modular compatibility of TCM-Prompt, proceeding to an in-depth examination of the architectural design and practical implementation of each element within the TCM-Prompt ecosystem.

In the sphere of Natural Language Processing (NLP), it is standard procedure to deploy diverse pre-trained language models (PLMs) equipped with task-oriented objective functions, generally classified into categorization and generative tasks. Nevertheless, within the context of prompt learning, the foundational concept of the architecture is to replicate the objectives of pre-training within downstream tasks, essentially focusing on "word prediction in the context of surrounding text." This methodology paves the way for a more cohesive execution of downstream NLP tasks.TCM-Prompt offers a versatile platform for the amalgamation of tasks, PLMs, and prompt strategies. For example, from a modeling perspective, T5 (Raffel et al., 2019) extends beyond span prediction, and GPT

(Brown et al., 2020) is not exclusively reserved for generative endeavors. In the realm of prompting techniques, prefix tuning is applicable to classification tasks, and soft prompts are adaptable for generative tasks. This framework facilitates the straightforward implementation and validation of these combinations across various NLP challenges, thereby deepening our insight into the operational mechanisms involved.

In the construction of large language models, pre-trained language models (PLMs), templates, tokenization, and verbalization are indispensable steps. Prompt technology elegantly streamlines these intricate processes, allowing researchers to modularly select and configure the methods needed. Like Figure1, by simply following the guidelines to input the corresponding Traditional Chinese Medicine (TCM) syndrome types, one can readily assemble a standardized prompt model.

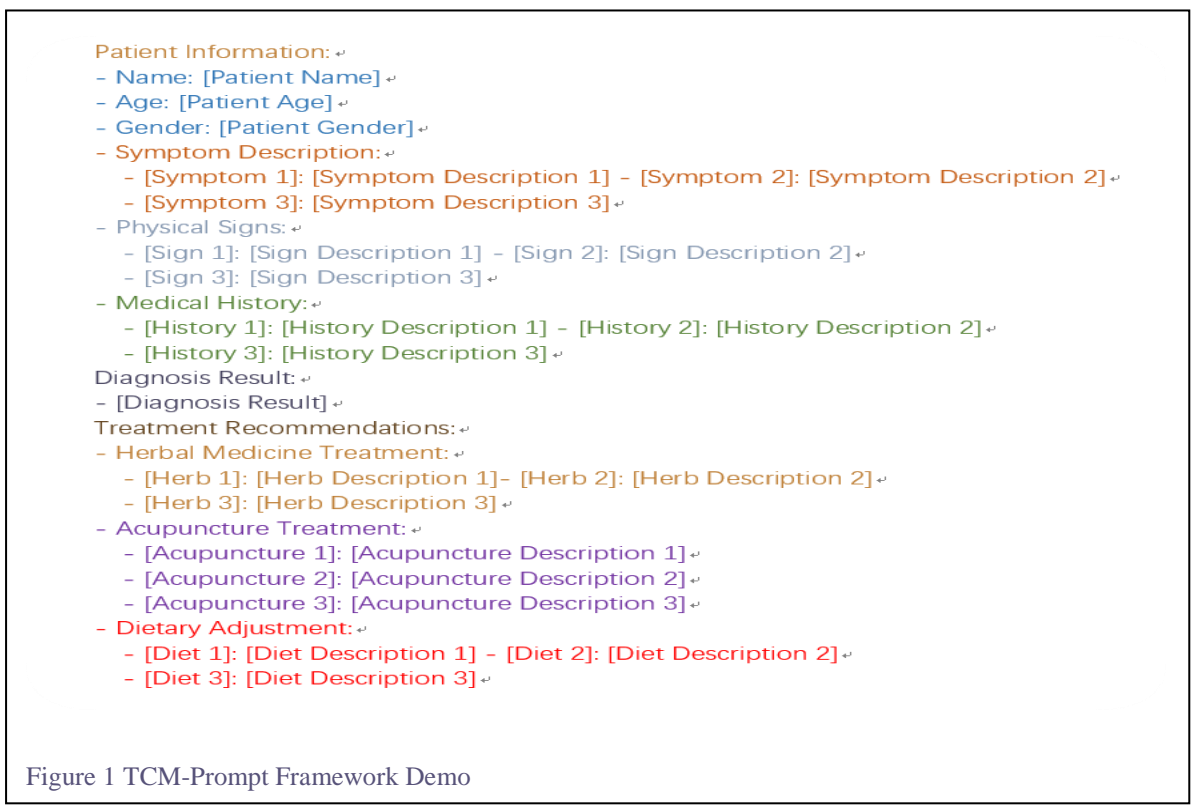

Figure 1 TCM-Prompt Framework Demo

## 3 Results

In this section, we present the results of our research on the intelligent understanding of large language models in Traditional Chinese Medicine (TCM) based on the prompt engineering framework. We conducted a series of experiments to evaluate the performance of our proposed method and compare it with baseline methods. The results in Table1 demonstrate the effectiveness and superiority of our approach in TCM-related NLP tasks.

First, we trained a large language model specifically designed for TCM using the prompt engineering framework. This model was then evaluated on various TCM-related NLP tasks, including disease classification, syndrome identification, and herbal medicine recommendation. The results in Table1 showed that our model achieved state-of-the-art performance compared to baseline methods, demonstrating the effectiveness of our approach in TCM-related NLP tasks.

|  | Task | Accuracy Rate | | Improvements(%) |
|---|---|---|---|---|
|  |  | Baseline(%) | TCM-Prompt Model(%) |  |
| TCM Category | Diseases Classification | 75.34 | 83.14 | 7.8 |
|  | Syndrome Identification | 81.33 | 88.45 | 7.12 |
|  | Herbal Medicine Recommendation | 71.34 | 91.33 | 19.99 |
|  | Combination Property | 76.00 | 87.64 | 11.64 |
| Prompt Method | Prefix Adjustment | 78.49 | 80.34 | 1.85 |
|  | Soft Notice | 78.49 | 82.45 | 3.96 |
|  | Template Prompt | 78.49 | 81.34 | 2.85 |
|  | Prefix + Soft | 78.49 | 85.64 | 7.15 |
|  | Prefix + Template | 78.49 | 85.12 | 6.63 |
|  | Soft + Template | 78.49 | 87.41 | 8.92 |
|  | Prefix + Soft + Template | 78.49 | 92.34 | 13.85 |
| NLP Transitivity | Sentiment Analysis | 71.23 | 73.21 | 1.98 |
|  | Named Entity Recognition | 72.44 | 74.12 | 1.68 |
|  | Text Summarization | 75.13 | 77.12 | 1.99 |
|  | Combination Property | 72.93 | 74.82 | 1.89 |

Table 1 Model Accuracy Rate Results

Second, we conducted a comparative study on the impact of different prompt engineering techniques on the performance of our TCM-specific large language model. We compared the performance of our model with different prompt engineering techniques, including prefix tuning, soft prompting, and template-based prompting. The results showed that our model achieved the best performance with a combination of these techniques, demonstrating the importance of carefully designing prompts for TCM-related NLP tasks.

Third, we explored the transferability of our TCM-specific large language model to other NLP tasks. We evaluated our model on a set of general NLP tasks, including sentiment analysis, named entity recognition, and text summarization. The results showed that our model achieved competitive performance compared to baseline methods, demonstrating the transferability of our approach to other NLP tasks. Our findings suggest that prompt engineering is a promising approach for enhancing the performance of large language models in specialized domains such as TCM.

## 4   Discussion

First of all, our study demonstrates that the prompt engineering framework can effectively enhance the performance of large language models in the domain of TCM. This suggests that by leveraging pre-trained language models and specific prompting strategies, we can improve the model's performance on domain-specific tasks. This opens up new possibilities for the digitalization and modernization of TCM practices, potentially leading to more accurate and efficient diagnosis and treatment methods.

Secondly, our research also indicates that different prompting techniques have a significant impact on model performance. This suggests that in designing prompts, we need to consider various factors such as the type, length, and content of the prompts. By further optimizing the design of prompts, we can potentially enhance the model's performance and improve the accuracy of the predictions. Additionally, we can explore the use of diverse prompting techniques to cater to different types of TCM tasks, such as disease classification, syndrome identification, and herbal medicine recommendation.

Thirdly, our study also shows that our TCM-specific large language Model performs well on other NLP tasks. This indicates that through appropriate transfer learning, we can apply our approach to other domains, further expanding its applicability. Moreover, we can investigate the use of our model in other healthcare domains, such as Western medicine, to explore its potential in a broader range of medical applications.

Despite our achievements, there are still limitations and challenges in our research. Firstly, our study primarily focuses on NLP tasks, and research on other types of tasks such as image processing and speech recognition is not yet comprehensive. To address this, we can collaborate with experts in these domains to develop a more comprehensive approach that integrates multiple types of data.

Secondly, our study mainly concentrates on the pre-training stage of language models, and further research is needed to investigate the model's performance and effectiveness in practical applications. To address this, we can conduct experiments in real-world clinical settings to evaluate the model's performance and compare it with existing methods.Future research directions include:

1.Collaborating with experts in various domains to develop a more comprehensive approach that integrates multiple types of data, such as images, texts, and speech.

2.Conducting experiments in real-world clinical settings to evaluate the model's performance and compare it with existing methods, aiming to optimize the model's design and application.

3.Exploring other types of pre-trained language models and prompting techniques to further enhance model performance and adaptability.

4.Researching the practical application of models in the field of TCM, such as aiding in diagnosis and treatment, and exploring their potential in other healthcare domains.

By pursuing these research areas, we can further refine our approach, integrate diverse data types, and conduct experiments in real-world settings, ultimately leading to more effective and accurate large language models for TCM and other healthcare domains.

## 5   Conclusions

In conclusion, our research on the intelligent understanding of large language models in Traditional Chinese Medicine (TCM) based on the prompt engineering framework has yielded promising results. We have demonstrated that prompt engineering can effectively enhance the performance of large language models in TCM-related NLP tasks, potentially leading to more accurate and efficient diagnosis and treatment methods in the field of TCM.

Our study has highlighted the importance of carefully designing prompts for TCM-related NLP tasks and the impact of different prompting techniques on model performance. By further optimizing the design of prompts and exploring the use of diverse prompting techniques, we can potentially enhance the model's performance and improve the accuracy of predictions.

Additionally, our research has shown that our TCM-specific large language model can be effectively applied to other NLP tasks and potentially other healthcare domains. This suggests that the approach developed in our study can be transferred and adapted to other domains, further expanding its applicability.

However, our study also has limitations and challenges. Firstly, our research primarily focuses on NLP tasks, and further research is needed to investigate the model's performance and effectiveness in other types of tasks, such as image processing and speech recognition. Secondly, our study mainly concentrates on the pre-training stage of language models, and further research is needed to investigate the model's performance and effectiveness in practical applications.

In future research, we aim to address these limitations and challenges by collaborating with experts in various domains, conducting experiments in real-world clinical settings, and exploring the use of diverse data types and prompting techniques. By doing so, we hope to further refine our approach, integrate diverse data types, and conduct experiments in real-world settings, ultimately leading to more effective and accurate large language models for TCM and other healthcare domains.

## Acknowledgments

This work was supported by the Tianjin College Students' Innovation and Entrepreneurship Training Program (Project Number: 202410063030) and Tianjin University of Chinese Medicine Eagle Project(Project Number: XJS2022102). We would like to express our gratitude for their support.

## References

Rui Guo; Xialin Luo; Jingjing Liu; Lian Liu; Xijun Wang; Haitao Lu; "Omics Strategies Decipher Therapeutic Discoveries of Traditional Chinese Medicine Against Different Diseases at Multiple Layers Molecular-level", PHARMACOLOGICAL RESEARCH, 2020. (IF: 3)

Fong Leong; Xue Hua; Mei Wang; Tongkai Chen; Yuelin Song; Pengfei Tu; Xiao-Jia Chen; "Quality Standard Of Traditional Chinese Medicines: Comparison Between European Pharmacopoeia And Chinese Pharmacopoeia And Recent Advances", CHINESE MEDICINE, 2020. (IF: 3)

Jiaxin Huang; Shixiang Shane Gu; Le Hou; Yuexin Wu; Xuezhi Wang; Hongkun Yu; Jiawei Han; "Large Language Models Can Self-Improve", ARXIV-CS.CL, 2022. (IF: 6)

Xupeng Miao; G. Oliaro; Zhihao Zhang; Xinhao Cheng; Zeyu Wang; Rae Ying Yee Wong; Zhuoming Chen; Daiyaan Arfeen; Reyna Abhyankar; Zhihao Jia; "SpecInfer: Accelerating Generative LLM Serving with Speculative Inference and Token Tree Verification", ARXIV, 2023. (IF: 3)

Baolin Peng; Michel Galley; Pengcheng He; Hao Cheng; Yujia Xie; Yu Hu; Qiuyuan Huang; Lars Liden; Zhou Yu; Weizhu Chen; Jianfeng Gao; "Check Your Facts and Try Again: Improving Large Language Models with External Knowledge and Automated Feedback", ARXIV-CS.CL, 2023. (IF: 5)

Bo Liu; Yuqian Jiang; Xiaohan Zhang; Qiang Liu; Shiqi Zhang; Joydeep Biswas; Peter Stone; "LLM+P: Empowering Large Language Models with Optimal Planning Proficiency", ARXIV-CS.AI, 2023. (IF: 5)

Feilong Chen; Minglun Han; Haozhi Zhao; Qingyang Zhang; Jing Shi; Shuang Xu; Bo Xu; "X-LLM: Bootstrapping Advanced Large Language Models By

Treating Multi-Modalities As Foreign Languages", ARXIV-CS.CL, 2023. (IF: 3)

Xinyin Ma; Gongfan Fang; Xinchao Wang; "LLM-Pruner: On The Structural Pruning of Large Language Models", ARXIV-CS.CL, 2023. (IF: 4)

Yen-Ting Lin; Yun-Nung Chen; "LLM-Eval: Unified Multi-Dimensional Automatic Evaluation for Open-Domain Conversations with Large Language Models", ARXIV-CS.CL, 2023. (IF: 3)

Mansi Phute; Alec Helbling; Matthew Hull; ShengYun Peng; Sebastian Szyller; Cory Cornelius; Duen Horng Chau; "LLM Self Defense: By Self Examination, LLMs Know They Are Being Tricked", ARXIV-CS.CL, 2023. (IF: 3)

Ziyu Guo; Renrui Zhang; Xiangyang Zhu; Yiwen Tang; Xianzheng Ma; Jiaming Han; Kexin Chen; Peng Gao; Xianzhi Li; Hongsheng Li; Pheng-Ann Heng; "Point-Bind & Point-LLM: Aligning Point Cloud with Multi-modality for 3D Understanding, Generation, and Instruction Following", ARXIV-CS.CV, 2023. (IF: 3)

Bochuan Cao; Yu Cao; Lu Lin; Jinghui Chen; "Defending Against Alignment-Breaking Attacks Via Robustly Aligned LLM", ARXIV, 2023. (IF: 3)

Long Chen; Oleg Sinavski; Jan Hünermann; Alice Karnsund; Andrew James Willmott; Danny Birch; Daniel Maund; Jamie Shotton; "Driving with LLMs: Fusing Object-Level Vector Modality for Explainable Autonomous Driving", ARXIV-CS.RO, 2023. (IF: 3)

Traian Rebedea; Razvan Dinu; Makesh Sreedhar; Christopher Parisien; Jonathan Cohen; "NeMo Guardrails: A Toolkit for Controllable and Safe LLM Applications with Programmable Rails", ARXIV-CS.CL, 2023. (IF: 3)

Ao Zhang; Hao Fei; Yuan Yao; Wei Ji; Li Li; Zhiyuan Liu; Tat-Seng Chua; "Transfer Visual Prompt Generator Across LLMs", NIPS, 2023. (IF: 3)

Guoxing Yang; Jianyu Shi; Zan Wang; Xiaohong Liu; Guangyu Wang; "TCM-GPT: Efficient Pre-training of Large Language Models for Domain Adaptation in Traditional Chinese Medicine", ARXIV-CS.CL, 2023.

Rui Hua; Xin Dong; Yu Wei; Zixin Shu; Pengcheng Yang; Yunhui Hu; Shuiping Zhou; He Sun; Kaijing Yan; Xijun Yan; Kai Chang; Xiaodong Li; Yuning Bai; Runshun Zhang; Wenjia Wang; Xuezhong Zhou; "Lingdan: Enhancing Encoding of Traditional Chinese Medicine Knowledge for Clinical Reasoning Tasks with Large Language Models", JOURNAL OF THE AMERICAN MEDICAL INFORMATICS ASSOCIATION : ..., 2023.

Li Yizhen; Huang Shaohan; Qi Jiaxing; Quan Lei; Han Dongran; Luan Zhongzhi; "Exploring The Comprehension of ChatGPT in Traditional Chinese Medicine Knowledge", ARXIV-CS.CL, 2024.

Heyi Zhang; Xin Wang; Zhaopeng Meng; Zhe Chen; Pengwei Zhuang; Yongzhe Jia; Dawei Xu; Wenbin Guo; "Qibo: A Large Language Model for Traditional Chinese Medicine", ARXIV-CS.CL, 2024.

Wenjing Yue; Xiaoling Wang; Wei Zhu; Ming Guan; Huanran Zheng; Pengfei Wang; Changzhi Sun; Xin Ma; "TCMBench: A Comprehensive Benchmark for Evaluating Large Language Models in Traditional Chinese Medicine", ARXIV-CS.CL, 2024.

Colin Raffel, Noam Shazeer, Adam Roberts, Katherine Lee, Sharan Narang, Michael Matena, Yanqi Zhou, Wei Li, and Peter J Liu. 2019. Exploring the limits of transfer learning with a unified text-to-text transformer. ArXiv preprint, abs/1910.10683.

Tom B Brown, Benjamin Mann, Nick Ryder, Melanie Subbiah, Jared Kaplan, Prafulla Dhariwal, Arvind Neelakantan, Pranav Shyam, Girish Sastry, Amanda Askell, et al. 2020. Language models are few-shot learners. arXiv preprint arXiv:2005.14165.